\documentclass[final,5p,times,twocolumn,authoryear]{elsarticle}

\usepackage{amssymb}
\usepackage{amsmath,amssymb,amsfonts,multirow}

\journal{Engineering Applications of Artificial Intelligence}

\begin{document}

\begin{frontmatter}



\title{M-FasterSeg: An Efficient Semantic Segmentation Network Based on Neural Architecture Search}

\tnotetext[t1]{This work was supported in part by the Key Area Research Projects of Universities of Guangdong Province under Grant 2019KZDZX1026,in part by the Natural Science Foundation of Guangdong under Grant 2020A1515110255, in part by the Science and Technology Program of Guangzhou under Grant 202102080591, in part by the Innovation Team Project of Universities of Guangdong Province under Grant 2020KCXTD015.}

\author[mymainaddress]{Junjun Wu}
\ead{jjunwu@fosu.edu.cn}


\author[mymainaddress]{Huiyu Kuang}
\ead{2111851017@stu.fosu.edu.cn}

\author[mymainaddress]{Qinghua Lu \corref{mycorrespondingauthor}}
\cortext[mycorrespondingauthor]{Corresponding author}
\ead{qhlu@fosu.edu.cn}


\author[mymainaddress]{Zeqin Lin}
\ead{linzeqin@fosu.edu.cn}

\author[mymainaddress]{Qingwu Shi}
\ead{2111951004@stu.fosu.edu.cn}

\author[mymainaddress]{Xilin Liu}
\ead{2112052028@stu.fosu.edu.cn}

\author[mymainaddress]{Xiaoman Zhu}
\ead{2111951003@stu.fosu.edu.cn}

\address[mymainaddress]{School of Mechatronic Engineering and Automation, Foshan University, Foshan, China}






\begin{abstract}
Image semantic segmentation is one of the key technologies for intelligent systems to understand natural scenes. As one of the important research directions in the field of visual intelligence, this technology has a wide range of application scenarios in the fields of mobile robots, drones, and intelligent driving. However, in practical applications, there may be problems such as inaccurate prediction of semantic labels, loss of segmented objects and background edge information. This paper proposes an improved semantic segmentation network that combines self-attention module and neural architecture search (NAS) method. The method first uses the NAS method to find a semantic segmentation network with multiple resolution branches. During the search process, the searched network structure is adjusted by combining the self-attention module, and then combined with the semantic segmentation networks searched by different branches to integrate into two semantic segmentation network models with different complexity, and finally integrate two network models with different complexity according to the current general teacher-student framework. The input image will first pass through the high complexity model to obtain more accurate parameters, which will affect the training weight of the student network, then pass the image into the low-complexity model to get the final predicted result. The experimental results on the Cityscapes dataset show that the accuracy of the algorithm is 69.8 \%, the inference speed is 166.4 FPS, and the actual image segmentation speed is 48/s. It can optimize edge segmentation for better performance in complex scenes and achieve a good balance between real-time performance and accuracy in practical applications.
\end{abstract}


\end{frontmatter}


\section{Introduction}
\label{sec:introduction}
Image  semantic segmentation is a technology that can automatically segment images and recognize different types of objects through a computer. At this stage, it is mainly studied by deep learning methods. The semantic segmentation network model based on deep learning neural networks has entered a period of vigorous development starting from Fully Convolutional neural Networks (FCN) \citep{r1}. However, the FCN method has two disadvantages: loss of pixel position information and lack of image context information. Therefore, various methods of optimizing convolution structure and feature fusion methods based on encoder-decoder network are commonly used in artificially designed semantic segmentation network models based on deep learning.

The artificially designed semantic segmentation network structure has a positive effect on the development of semantic segmentation. For example, the introduction of dilated convolution in DeepLabv1 optimizes the resolution information of the feature map \citep{r2}. DeepLabv2 introduces the combination of the spatial pyramid pooling model and the dilated convolutional layer to increase the ability to acquire contextual semantic information \citep{r3}. DeepLabv3 proposes a combination of multiple combinations and multiple dilated convolution layers to capture multi-scale context information\citep{r4}. The encoder-decoder structure is adopted in DeepLabv3+ to improve the context information acquisition capability\citep{r5} Although this type of artificially designed semantic segmentation network model can effectively improve the recognition accuracy of the semantic segmentation model, it requires a lot of time-consuming manual design time and computing resources, and it is difficult to apply it to practical applications.

Based on the problem that the artificially designed semantic segmentation network takes too long and consumes too much computing resources, we uses neural network architecture search technology to make the computer automatically search to generate the semantic segmentation network model. Based on the DeepLab series, we propose a semantic segmentation network model that combines attention mechanism and neural network architecture search technology. This method can optimize image output branches of different resolutions, use horizontal/vertical domain information to model various types of objects in the picture, and introduce global contextual semantic information to remove the feature information of non-recognized objects. From the local to the global two learning stages, more accurate image semantic feature information can be learned. The neural network search technology is introduced to reduce the time spent in artificially designing the neural network structure, and the knowledge distillation network is used to obtain a neural network structure model for fast semantic segmentation. The innovation points are as follows.

\begin{enumerate}
\item {A novel attention network module that can be applied in the field of real-time semantic segmentation.}
\item{optimize the semantic segmentation network of NAS search and use the knowledge distillation network to expand it. It is proposed to use a new optimization function to achieve rapid convergence for semantic segmentation and shorten the training time.}
\end{enumerate}

\section{Related Work}
The neural network structure model based on neural network search technology has been successful in the field of semantic segmentation. Researchers have proposed a variety of ways to use neural network search technology to improve the accuracy of image semantic segmentation tasks. Related to the work of this paper are key technologies such as fast neural network search technology, adaptive attention mechanism and knowledge distillation network.

\subsection{Fast neural network architecture search technology}
The fast neural network architecture search technology comes from the Neural Architecture Search (NAS) \citep{r6}. NAS technology uses reinforcement learning \citep{r7}, evolutionary algorithm \citep{r8} and gradient algorithm \cite{r9} three basic construction methods to set the search space, and the cellular network construction method is used to set the basic neural network component modules as the cellular network structural blocks. Cell network structure is combined into a new neural network structure model \citep{r10}. Compared with the artificial design of neural network structure module, this method saves time and effort and is completely designed by computer training, which is convenient to find the optimal network structure.

Based on the above concepts, Chen et al. applied NAS technology to the field of semantic segmentation. At this time, the semantic segmentation network model designed based on NAS also needs to consume huge computing resources \citep{r11}. Li Feifei and others designed Auto-DeepLab network structure model based on the concept of DeepLab series dilated convolution \citep{r12}. Although this network model greatly reduces the consumption of computing resources, it still requires three GPUs to work at the same time for a week to complete. Chen et al. proposed a hybrid multi-resolution branch structure to speed up the search speed of the neural network architecture. This method can be performed on a single GPU, but it still takes nearly a month to complete the training \citep{r13}. 

\subsection{Attention mechanism module}
The design inspiration of the attention model is to solve the pixel loss problem caused by the importance of different channels of the feature map in the process of convolutional layer pooling in deep learning. For semantic segmentation tasks, it is necessary to obtain precise pixel positions in special images to obtain better semantic segmentation results. Based on this, Jung et al. proposed DANet two-way attention scene image segmentation network model. The network structure model uses the dual attention module to integrate the global semantic information of the image into the network, and the output results pass through the two attention modules and then pass through the convolutional layer. Element weighting realizes feature fusion, and the output result is connected to the convolutional layer to obtain the final prediction result \citep{r14}. Hu et al. proposed that ACNet combined with the attention network model to extract object image depth information can achieve good results in the semantic segmentation of indoor scenes \citep{r15}. This type of neural network model with attention network is very effective in improving the accuracy of image segmentation, but it also has common shortcomings that will slow down the model training speed. Based on the attention module theory, we proposes a two-round four-direction attention model that can simultaneously obtain horizontal/vertical information, expand the receptive field and achieve the effect of rapid recognition and segmentation. At the same time, the residual network module is added to extract the global information of the pixel, and the global semantic information is integrated to improve the recognition accuracy.

\subsection{Knowledge distillation module}
There are two ways to achieve better prediction results in deep learning: using regularization functions and a wide and deep network structure as neural network model training and using a variety of weak model sets. Both of the above two methods can achieve better prediction results, but the disadvantage lies in the huge consumption of computing resources. Therefore, Hinton first proposed the concept of knowledge distillation in the article. In the article, he believes that the concepts of teacher network and student network can be introduced to reduce the consumption of computing resources. The teacher network refers to a complex but superior performance network structure, and the student network refers to a simple, low-complexity, and fast-training network structure. Use knowledge transfer to use the training weights in the teacher network to guide the training of the student network, thereby reducing the training time of the student network\citep{r16}. In the field of semantic segmentation, Liu et al. introduced the concept of knowledge distillation into the field of semantic segmentation for the first time \citep{r17}. We use knowledge distillation to achieve rapid semantic segmentation.

\section{The Proposed Semantic Segmentation Network}
\subsection{Adaptive spatial attention network module}
In response to the current situation of the problems described in the previous chapters, this chapter proposes a NAS-based fast semantic segmentation network model (More FasterSeg, M-FasterSeg), which improves the delay response performance while pursuing the accuracy of semantic segmentation. The core components of the overall structure of the fast semantic segmentation network model M-FasterSeg proposed in this study include: adaptive attention mechanism, neural network architecture search and knowledge distillation. As shown in Figure 1, the workflow of the network model is described as follows: When an image is input to M-FasterSeg, it first passes through the neural architecture search module, which uses super network technology combines the adaptive attention mechanism module to generate two different network structure models based on the input image: Teacher network model (Teacher) and student network model (Student); then use knowledge distillation to combine the teacher network and student network; finally the image is processed by the two neural network structures in the knowledge distillation network and then output the final semantic segmentation map.

\begin{figure}[!t]
\centering
\includegraphics[width=\columnwidth]{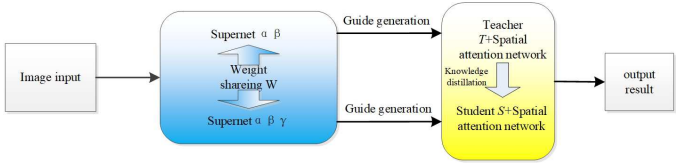}
\caption{Quick semantic segmentation algorithm Chart.}
\label{fig1}
\end{figure}

\subsubsection{M- FasterSeg adaptive attention module}
In order to enhance the accuracy of semantic segmentation and the low-latency response performance of the network model, this section combines the IRNN network model  \citep{r18}and the attention mechanism \citep{r19} to propose a new  adaptive attention mechanism module. Among them, the IRNN network model is often used in the field of natural language processing. It is composed of a ReLU activation function and a recurrent neural network initialized by the unit matrix. When performing natural language processing, it shows good characteristics of high accuracy and low latency; attention mechanism can remove the invalid feature channel information, and retain more effective feature channel information, then the working accuracy of the network model can be improved.

The spatial attention network module structure designed in this paper, as shown in Figure 2, uses two sets of IRNN network structures as the basic structure for obtaining semantic information. On this basis, additional convolutional layer branches are added to capture spatial context information. In this branch, the super-supervised method uses the convolutional layer and the sigmoid activation function module to generate the attention map, and the physical distribution of the semantic segmentation object is displayed in the attention map to guide the subsequent output process of the semantic segmentation result map. Using the spatial attention module can effectively identify target objects in complex environments and similar target objects with changing appearances, even in the case of multi-target aliasing.

\begin{figure*}[!t]
\centering
\includegraphics[width=18cm]{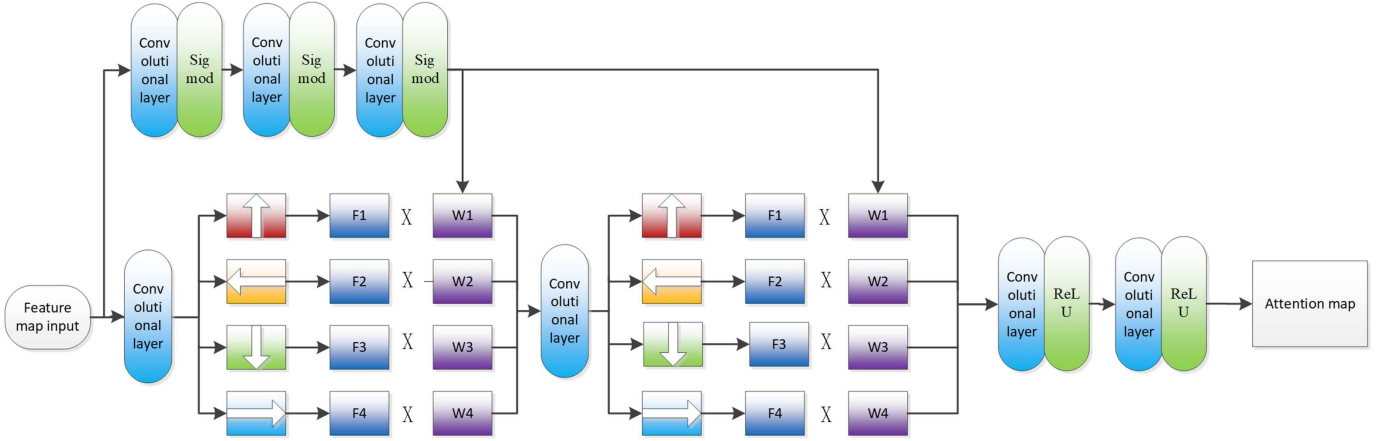}
\caption{Spatial attention module.}
\label{fig2}
\end{figure*}

\subsubsection{Adaptive spatial attention network module}
The spatial attention module designed in this paper uses dimensional compression and information activation methods in the convolutional layer branch and the attention map output part to distinguish the importance of each feature channel output, and discard the invalid feature channel information according to its importance. Enhance effective feature channel information, so that the network model can obtain more pixel information and improve the accuracy of the model.

The dimensional compression method can make the convolution kernel obtain enough information in the same local space to extract the relationship between each feature channel. This method uses the global average pooling layer to encode the spatial feature on a single feature channel into a global feature descriptor. The specific process description is shown in formula (1):
\begin{equation}
z_{c}=F_{s q}\left(u_{c}\right)=\frac{1}{H \times W} \sum_{i=1}^{H} \sum_{j=1}^{W} u_{c}(i, j)
\end{equation}

In the formula, H and W respectively represent the length and width of the feature map of the input attention module, $u_c(i,j)$  refers to the output value of the convolution kernel at the pixel position $(i,j)$, and $u_c$ refers to the output of the c-th convolution kernel.

After dimensional compression, the information activation method can enable the attention module to divide the importance of the relationship between the feature channels by virtue of the global feature descriptor. Using a gate mechanism with a sigmoid form can enable the attention module to allow multiple feature channel information to be input at the same time when learning the nonlinear relationship between each feature channel. The learning process can be described by the formula (2):
\begin{equation}
s=F_{e x}(z, W)=\sigma(g(z, W))=\sigma\left(W_{2} R e L U W_{1} z\right)
\end{equation}

In the formula, $W_1$ is the dimension of the subset after the feature map is compressed, and $W_2$ is the dimension of the subset after the feature map is restored.
After the information activation method, the size of the feature map becomes smaller, and the ReLU activation function needs to be used to restore the feature map to the original size. Finally, the effective feature channel signal activation value is multiplied with the original feature value output by the convolution kernel to strengthen the effective feature information channel. The process description is shown in formula (3):
\begin{equation}
x c=\operatorname{Fscale}\left(u_{c}, s_{c}\right)=s_{c} \cdot u_{c}
\end{equation}

In the formula, $s_c$ is the output result of the feature channel learned by the fully connected layer in the attention module, and $u_c$ is the output result of the convolution kernel in the attention module.

In the spatial attention module designed in this paper, the attention module is used as the basis, and on this basis, two IRNN network models are used to form a two-round four-way IRNN network model structure to improve the information discrimination ability of the attention module. After the feature map information is input to the spatial attention network module, the feature map is processed using the two-round four-way IRNN model, and the processed output result is multiplied by the output result of the information activation method in the attention branch, and the feature map size is restored through ReLU activation, and then get the attention map of the original size.

The IRNN network model consists of four direction-independent RNN networks, and the model structure is shown in Figure 3. The IRNN network divides the convolution operations in four different directions (up, down, left, and right) into independent convolutional layers. The weight information obtained by the convolutional layers in different directions is shared with each other. All operations use recursive matrices and nonlinearities activation function to transfers information, and calculates by the pixel feature information obtained from each spatial position in the four directions. The IRNN module $h_{(i,j)}$ moving to the right updates the calculation formula at the position $(i,j)$, such as the formula (4) Shown:
\begin{equation}
h_{i, j}^{r i g h t} \leftarrow \max \left(h_{i, j-1}^{\text {right }}+h_{i, j}^{\text {right }}, 0\right)
\end{equation}

In the formula, $h_{(i,j)}^{right}$ represents the value of the IRNN module moving to the right to obtain the value of maximum pixel information. The IRNN module moving to the right can combine the previous image position information and the current image position information to calculate and retain the maximum value information. The same goes for other directions. Therefore, the IRNN module can effectively obtain larger receptive field information, thereby effectively enhancing the model's ability to discriminate feature information in the feature map.

\begin{figure}[!t]
\centering
\includegraphics[width=4cm]{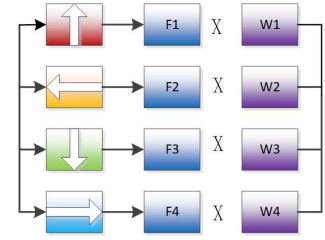}
\caption{IRNN network structure model.}
\label{fig3}
\end{figure}

In the spatial attention module designed in this article, the two-round four-direction IRNN network structure model is used, which not only enables the spatial attention module to obtain a larger receptive field during image training, but also after the output feature map of the first set of IRNN network structure models is input to the second set of IRNN network structure models, the second set of IRNN network structure models can combine the feature information of the re-input feature map and the original input feature map for judgment, so as to obtain more contextual semantic information, which is more conducive to semantic segmentation task, the working principle diagram of two groups of IRNN structure, as shown in Figure 4. In Figure 4 (a), IRNN performs a four-way (up, down, left, and right) convolution operation on each pixel position of the input feature map to collect horizontal /neighborhood information. In the second stage, as shown in Figure 4 (b), the context information is obtained from the re-input feature map by repeating the steps of the previous stage, and finally the complete feature map is output.

\begin{figure}[!t]
\centering
\includegraphics[width=\columnwidth]{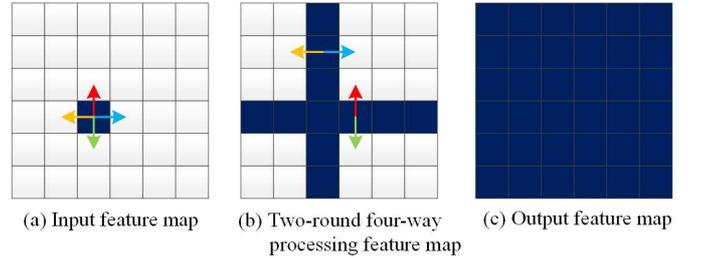}
\caption{IRNN working principle diagram.}
\label{fig4}
\end{figure}

In order to retain more image feature information and reduce the model delay time, the method in this paper uses the spatial attention module as the main body to extract the object image information, and builds an adaptive attention network module, as shown in Figure 5. The adaptive attention network module adopts the residual network structure and jump connection to realize the retention of more image feature information and reduce the delay time. The residual network can be used to establish identity mapping in the input and output of the feature map, the original feature map information is added with the feature information in the attention map output by the spatial attention network module. After the activation function, the image size is restored to the pixel feature result map with the same size as the input image, and the jump connection can be used to suppress the degeneration of neural network, to achieve the effect of accelerating the speed of convergence. Not only that, after using jump connections, the path of data input can be shortened and the spread of false information can be suppressed, so that the adaptive attention module can retain more correct image semantic information. The experiment in Chapter 4 verifies this fact: the use of skip connection and residual module in the adaptive attention module can effectively improve the accuracy of the model's semantic segmentation and reduce the delay time.

\begin{figure}[!t]
\centering
\includegraphics[width=4cm]{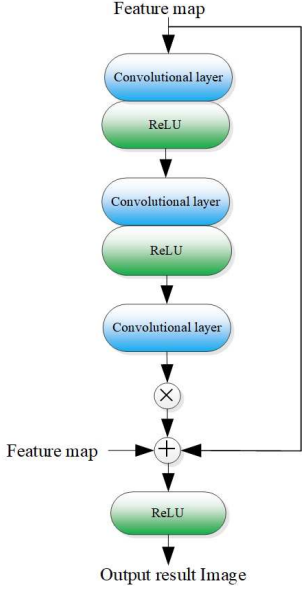}
\caption{Adaptive spatial attention network module.}
\label{fig5}
\end{figure}

\subsection{M- FasterSeg neural network architecture search module}
In order to reduce the time consumed by artificial neural network design, this paper automatically searches and generates semantic segmentation network models on computing workstations based on neural network architecture search technology. This paper proposes a new neural network architecture search technology, which can effectively improve the accuracy and delay time of automatically generating semantic segmentation neural network models. This technology is based on NAS technology  \citep{r20} combined with an adaptive attention network module to process multiple resolution branches, and finally merge the output of multiple resolution branches to generate the final semantic segmentation network model.

The structure diagram of the neural network architecture search module designed in this paper is shown in Figure 6. First, the image enters the NAS backbone network, and the cell network uses reinforcement learning to search, and then the image is resized to generate feature map information with different resolution branches, and then the adaptive spatial attention network module is used to distinguish feature information in different feature channels, then remove the invalid feature channel information, and finally the outputs of different resolution branches are merged to generate a new semantic segmentation network model.

\begin{figure*}[!t]
\centering
\includegraphics[width=16cm]{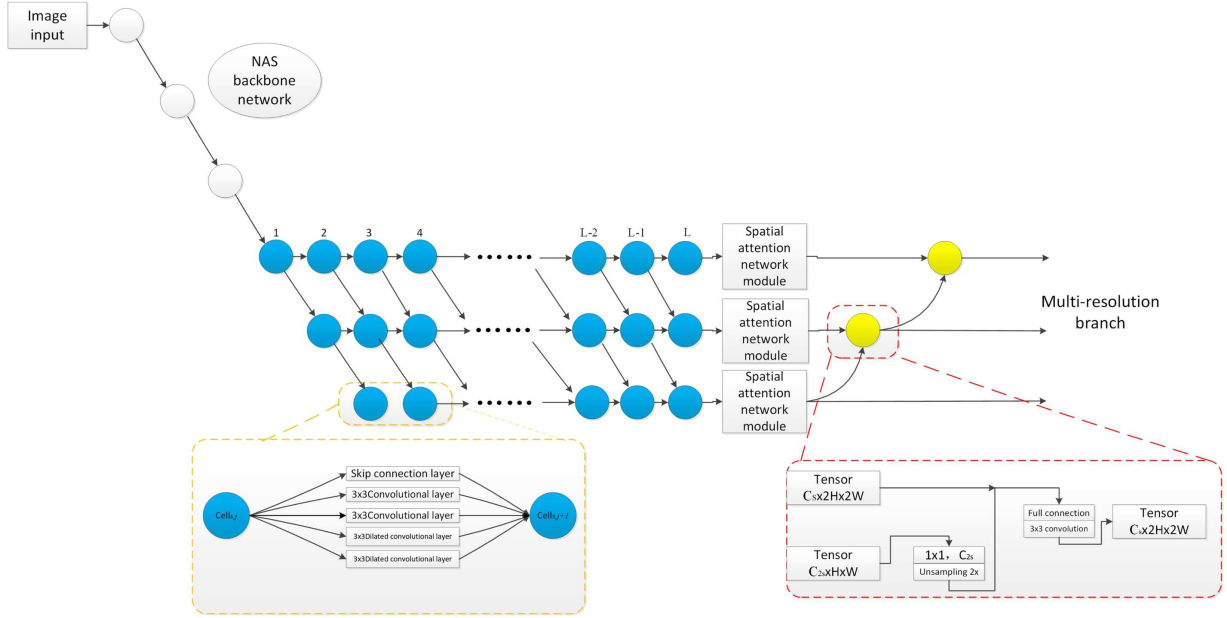}
\caption{The search space structure diagram of adding an adaptive spatial attention network module to the multi-resolution.}
\label{fig6}
\end{figure*}

Search module proposed in this paper adopts the method of continuous expansion of the search space in the search process. After the image enters the structure search process, the search technology will search the topology in the network to obtain more image semantic information, thereby improving the accuracy of the generated semantic segmentation model. The search space continuous expansion method can not only search the topology of the network, but also search the width of the network and the structure of the convolution unit. The continuous expansion of search space method  adopted in this paper can connect each unit with two possible previous search unit outputs, thereby improving the accuracy of the generated semantic segmentation model. The formula is shown in (5):
\begin{equation}
\bar{O}_{s \rightarrow s, l}=\beta_{s, l}^{0} \bar{O}_{\frac{s}{2} \rightarrow s, l-1}+\beta_{S, l}^{1} \bar{O}_{s \rightarrow s, l-1}
\end{equation}

In the formula, s represents the down-sampling rate, the layer index is $l$, $\beta$ is a constant value, each operator $O^k\in O$ and each pre-output is set to $\bar{O}_{(l-1)}$. In the space expansion method, each search unit can have two outputs, which are output to the subsequent units at different down-sampling rates. The expression formulas are as shown in (6) and (7):
\begin{equation}
\bar{O}_{s \rightarrow s, l}=\sum_{k=1}^{|o|} \alpha_{s, l}^{k} O_{s \rightarrow s, l}^{k}\left(\bar{I}_{s, l}, x_{s, l}^{j}, \text { stride }=1\right)
\end{equation}
\begin{equation}
\bar{O}_{s \rightarrow 2 s, l}=\sum_{k=1}^{|O|} \alpha_{s, l}^{k} O_{s \rightarrow 2 s, l}^{k}\left(\bar{I}_{s, l}, x_{s, l}^{j}, \text { stride }=2\right)
\end{equation}

In the formula, $x_{(s,l)}^j$ is the coefficient of sampling value of the random ratio in $p = x=x_{(s,l)}^j= \gamma_{(s,l)}^j$ using the Gumbel-softmax method, in the formula, $\alpha, \beta, \gamma$ is used as a normalized scalar to calibrate each unit output operator.
Since the Gumbel-softmax method can model the output of the search unit in the architecture search technology as a discrete multinomial distribution and derivable mathematical model, thereby generating a semantic segmentation network method, the Gumbel noise value can be set as shown in formula (9), and the output result is shown in formula (9), in which the parameter $\tau=1$.
\begin{equation}
o^{k}=-\log (-\log (u)),(u=0,1)
\end{equation}
\begin{equation}
j=\operatorname{argmax}\left(\frac{\exp \left(\frac{\log \left(\gamma^{i}+\sigma^{i}\right)}{\tau}\right)}{\sum_{m=1}^{|X|} \exp \left(\frac{\log \left(\gamma^{m}+\sigma^{m}\right)}{\tau}\right)}\right)
\end{equation}

This paper uses the spatial continuous expansion method to process each arithmetic unit of the neural network architecture, in which $\alpha,\beta,\gamma$ exists as a normalized scalar in the softmax layer, so as to realize the spatial continuous expansion method. All the normalized scalars are respectively related to each operator $O^k\in O$ the output of each pre-operator $\bar{O}_{(l-1)}$, and the calculation process of normalized scalar with expansion ratio of each expansion convolution is $x \in X$, as shown in formulas (10),(11),(12):
\begin{equation}
\sum_{k=1}^{|O|} a_{s, l}^{k}=1, \forall s, l\left(a_{s, l}^{k} \geq 0, \forall k, s, l\right)
\end{equation}
\begin{equation}
\beta_{s, l}^{0}+\beta_{s, l}^{1}=1, \forall s, l\left(\beta_{s, l}^{0}, \beta_{s, l}^{1} \geq 0, \forall k, s, l\right.
\end{equation}
\begin{equation}
\sum_{j=1}^{|x|} \gamma_{s, l}^{j}=1, \forall s, l\left(\gamma_{s, l}^{k} \geq 0, \forall j, s, l\right)
\end{equation}

Where $s$ is the down-sampling rate in the network model structure, and $l$ is the index of the number of network layers in the super network technology.
In the neural network architecture search technology designed in this paper, super network technology is used to generate two semantic segmentation network models. With supernet  technology, the architecture search technology can generate two semantic segmentation network models with different parameters at the same time with only one search training. After the picture is input into the architecture search technology designed in this article, two network models with different parameters are generated at the same time, and the knowledge distillation module is constructed with two network models with different parameters. Through knowledge distillation technology, the network model with a few parameters is used as the algorithm model of the fast semantic segmentation method, so as to realize the fast semantic segmentation.

In the current commonly used architecture search technology, a search process can only generate a network model of one parameter type. Therefore, adding super network technology to the architecture search technology proposed in this article can save training time, thus, two kinds of fast semantic segmentation network models with different complexity can be generated faster The advantage of super network technology is that the weight of the network model searched by NAS technology can be shared. However, when the commonly used super network technology is used in the search process, it will cause the search technology to fall into the "local minimum" when the search technology training converges, resulting in the poor accuracy of generated network structure, and because too many hop connections are often used for information transmission in the hyper-network  technology, this method causes the structure to collapse, and also causes the accuracy of the generated network structure model to decrease.

In this study, combined with the theory of super networks, it is found that the reason for the collapse of the architecture is that the super network has different sensitivity to the operator $O$, down-sampling rate $s$, and expansion rate $x$. In the separated convolution and the output of the convolution layer, a large delay gap will be shown, and the gap will not be too obvious in different resolution branches. Therefore, the regularized delay method is used to optimize the network search space of three different influencing factors. The optimization process is shown in (13):
\begin{equation}
\begin{aligned}
\text { Latency }(0, \mathrm{~S}, \mathrm{X})=&w_{1} \operatorname{Latency}(O \mid s, X)
+w_{2} \operatorname{Latency}\\&(s \mid 0, x) 
+w_{3} \text { Latency }(x \mid 0, s)
\end{aligned}
\end{equation}

Among them, $w_1,w_2, w_3$  are constants, $w_1=0.001,w_2=0.997, w_3=0.0021$, using separate influence factors for regularization optimization, which effectively reduces the problem of architecture collapse.

In the neural network architecture search technology designed in this article, the image will use the proposed adaptive attention network module to process the output feature map after the architecture search process and the aggregation branch process. This method can make the output branch feature information pass through the adaptive spatial attention network module, realize the recognition of the vertical/neighborhood information of the image feature information, combine the contextual semantic information and remove the invalid feature information, thereby improving the accuracy of the semantic segmentation network architecture model generated by the neural network architecture search technology designed in this paper.

\subsection{M-FasterSeg knowledge distillation module}
The neural network architecture search technology proposed in this article can generate two network models with different parameters in the same search process. The network model with unlimited parameters is called "teacher network", and a lightweight network model with limited parameters is called "student network". Among them, because the teacher network model needs to obtain more information, its search process will not be affected by the search expansion ratio, but will only search for the network structure with the widest expansion ratio, which leads to the unlimited number of parameters of the teacher model; When searching the student network structure, the search expansion ratio will be limited, so that the student network structure model will not be too large. The purpose of controlling the parameters of the student network structure model is to reduce the complexity of the student network and reduce the delay time of the student network model.

Using knowledge distillation technology  \citep{r21}, the teacher network structure can be combined with the student network. Under the same data set, the teacher network training weight is used as an impact factor to guide the student network training so as to improve the accuracy of the student network model. Therefore, this article uses knowledge distillation technology to achieve rapid semantic segmentation of the student network, and uses the adaptive attention module as a post-processing module in the teacher and student network training process to improve the verification accuracy of the teacher and student network model.

The knowledge distillation network method used in this article has the following two steps: (1) training the teacher network, (2) conveying the knowledge in the teacher network model to the student network structure model. In the training teacher network model, according to the Boltzmann distribution concept of statistical mechanics, the generalized softmax function with the temperature parameter T added is used, and the generalized softmax function formula with the temperature parameter is introduced as shown in (14):
\begin{equation}
q_{i}=\frac{\exp \left(z_{i} / \mathrm{T}\right)}{\sum_{j} \exp \left(z_{j} / T\right)}
\end{equation}

In the softmax evaluation of the pixel feature map, the output results of the pixel position $(i,j)$ are $z_j$ and $z_i$. According to this formula, the larger the value of the parameter T, the output probability in the softmax layer will gradually become smoother, resulting in a larger entropy value of the distribution, the information carried by the label will be amplified, and model training will pay more attention to the information of the label.

The second part is to transfer the image knowledge obtained by the teacher network to the student network structure. During the knowledge transfer process, the temperature parameter $T$ of the softmax layer in the student network needs to be restored to 1, so that the softmax layer pays more attention to the pixel information probability distribution. In the search process, teachers and students use the supernet to share the same network weight $W$, so the weight of the student network is the same as the weight of the teacher network, only need to apply the weight parameter obtained from the teacher network training as an impact factor to the student network training. Distillation process description, as shown in formula (15):
\begin{equation}
l_{\text {Distillation Loss }}=E_{i \in R} K L\left(q_{i}^{S} \| q_{i}^{t}\right)
\end{equation}

$KL$ is the divergence notation, $q_i^s$ and $q_i^t$ are the logarithms of the pixel predicted value i of the image in the teacher network and the student network, and the equal weight $(1,0)$ is assigned to the image segmentation loss and distillation network loss. After the search is completed to generate the teacher network and student network model, an adaptive attention network is added in the training process of the teacher network and the student network to improve the adaptive attention of the semantic segmentation training process, and the purpose is to improve the accuracy of semantic segmentation network and reduce delay time.

\section{Performance Verification of The Proposed Method}
{Figures that are meant to appear in color, or shades of black/gray. Such 
figures may include photographs, illustrations, multicolor graphs, and 
flowcharts.}
\subsection{Experimental dataset and test platform}
In this article, the Cityscapes dataset is used as the verification dataset. The image data in the Cityscapes dataset is mainly provided by three German companies such as Daimler. It includes street views in more than 50 cities in different environments, including 2975 training images. 500 pictures of the validation set and 1525 pictures of the test set. The Cityscapes data set collection scene is shown in Figure 7. 

\begin{figure}[!t]
\centering
\includegraphics[width=\columnwidth]{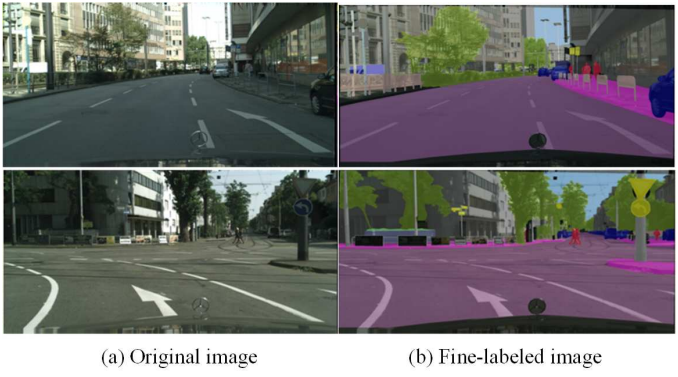}
\caption{The original image of the Cityscapes dataset and fine annotations.}
\label{fig7}
\end{figure}

Experimental test workstation hardware configuration conditions: NVIDIA Geforce GTX 1080ti GPU, Intel i7-8700kCPU and 16G memory. The computing performance of this platform is equivalent to the current mainstream mobile robot navigation control system platform configuration, so it can better evaluate the network model in the mobile robot platform Working performance. In order to quantitatively test the operating efficiency of the network model, the software configuration of the experimental platform is described as follows: TensorRT v6.0.1 deep learning inference framework is used to report the inference speed of each epoch during experimental training, Pytorchv1.4 is used as the deep learning framework, and the GPU uses CUDA10.1 and CUDNNV7 models

\subsection{Performance evaluation index}
\subsubsection{Accuracy evaluation index}
The experiment uses three indicators to quantitatively and qualitatively evaluate the accuracy performance of the proposed network model, including: pixel accuracy (Pixel Accuracy, PA), average pixel accuracy (mean Pixel Accuracy, mPA), average intersection and ratio (mean Intersection over Union, mIoU).
The "PA" indicator reflects the proportion of the correct pixels in the semantic segmentation result image to the total pixels in the image, and the calculation formula is shown in formula (16):
\begin{equation}
P A=\frac{\sum_{i=0}^{k} p_{i i}}{\sum_{i=0}^{k} \sum_{j=0}^{k} p_{i j}}
\end{equation}

The "mPA" indicator first calculates the proportion of pixels that are correctly classified in each type of pixel, and then calculates the average of the series of proportional values. The evaluation conditions for this indicator are more stringent than the indicator PA. The mPA calculation formula is shown in formula (17):
\begin{equation}
\mathrm{m} P A=\frac{1}{k+1} \sum_{i=0}^{k} \frac{p_{i i}}{\sum_{j=0}^{k} p_{i j}}
\end{equation}

The "mIoU" index is the most commonly used accuracy evaluation index in the field of semantic segmentation. In the semantic segmentation result map, the pixel set M of the real category object and the pixel set N of the correct category object are two important indicators for evaluating the semantic segmentation ability, and mIoU is the sum of the intersection and union of the sets M and N. The pixel sets of each category in the image are respectively calculated for the corresponding intersection and union sums, and the intersection and union sums of all the sets are averaged, and the evaluation result is finally obtained. The formula for calculating mIoU is shown in formula (18):
\begin{equation}
m I o U=\frac{1}{k+1} \sum_{i=0}^{k} \frac{p_{i i}}{\sum_{j=0}^{k} p_{i j}+\sum_{j=0}^{k} p_{j i}-p_{i i}}
\end{equation}

In Equation (16),(17),(18), $k+1$ is the number of categories in the data set. Generally, the category of the data set includes $k$ category labels and an unlabeled category. It is predicted to belong to the category of $j$, so $p_{ij}$, $p_{ii}$ and $p_{ji}$ respectively define the correct part, the wrong part and the wrong unknown category part.

\subsubsection{Real-time evaluation indicators}
The network model proposed in this paper is mainly oriented to the application of mobile robot navigation system, so it has higher requirements for computational efficiency. The experiment uses three indicators to evaluate the real-time performance of network model work: the number of floating point operations (FLOPs), the number of parameters (parameters) and the number of Frames Per Second (FPS), which are also commonly used in the field of fast semantic segmentation research. index. FLOPs reflect the speed of floating-point operations, which can be used to measure the computational complexity of the algorithm/model; Parameters reflect the size of the parameter memory when the network model is running. For a convolutional layer, the calculation methods of these two evaluation indicators are shown in formula (19) and formula (20):
\begin{equation}
\text { FLOPs }=\left(2 \times C_{i} \times K^{2}-1\right) \times \mathrm{H} \times \mathrm{W} \times C_{0}
\end{equation}
\begin{equation}
\text { Parameters }=\left(K^{2} \times C_{i}\right) \times C_{0}+C_{0}
\end{equation}

Where $C_i$ is the number of input channels in the convolutional layer, $K$ is the size of the convolution kernel, $H$ and $W$ are the sizes of the output feature maps, respectively, and $C_0$ is the number of output channels in the convolutional layer.

The FPS is used to reflect the image processing speed of the model and can be used to measure the image processing speed of the semantic segmentation model. The evaluation index calculation method is shown in formula (21):
\begin{equation}
{FPS}={F} \div {T}
\end{equation}

In the formula, $F$ is the frame number, and $T$ is the elapsed time, which is usually set to a constant of 1 second.

\subsection{Experiment Validation}
\subsubsection{Experimental program}
First, the Cityscapes data set is input into the neural network architecture search of M-FasterSeg. In the search technology training process, the supernet technology is used to generate the semantic segmentation network model for teachers and students at the same time. Secondly, for the teacher network model with adaptive attention post-processing module, Cityscapes is used as the training data set for training to obtain the final training weights. Finally, the knowledge distillation technology is used to input the training weights into the student network model as an impact factor. After the model, an adaptive attention module processing module is added, and Cityscapes is used as the training and test data set of the student network model to generate the result graph of the fast semantic segmentation method on the Cityscapes data set.

\subsubsection{Neural network architecture search performance evaluation }
This section mainly evaluates the accuracy and speed of the M-FasterSeg neural network architecture search technology to generate the network model. The network structure model in More FasterSeg proposed in this paper is searched by neural network architecture search technology combined with adaptive attention mechanism. In the search training process, it is based on the semantic segmentation proposed by Chen et al. (2019) \cite{r22} Multi-resolution branch neural network architecture search technical parameter setting, set the number of cell network layers $L=16$, the downsampling ratio $s\in \{8,16,32\}$, the number of multi-resolution branches is 2, and the downsampling of all downsampling layers The rate expansion ratio is $x_{(s,l)}\in k=\{4,6,8,10,12\}$. The Cityscapes dataset is used as the training and verification dataset for the search technology. The multi-resolution branch that reaches the appropriate depth fastest can share the data of the first three operators to another branch, and then diverge, and set the output branch of the network structure model to 2. Start training with $256\times512$ image size batch $size=3$ parameters.

The search technology proposed in this paper combined with the supernet method can simultaneously search on different output branches to generate teacher network and student network models. This paper conducts an experimental analysis on whether to add an adaptive attention module to the output branch of the search module to verify the effectiveness of the adaptive attention module designed in this paper. The evaluation indicators are accuracy and speed. The experimental results are shown in Figure 8 and As shown in Figure 9.
\begin{figure}[!t]
\centering
\includegraphics[width=\columnwidth]{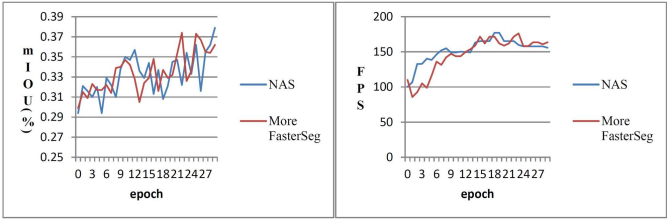}
\caption{Use the Cityscapes data set to use neural search technology to generate the teacher network in the knowledge distillation network.
On the left hand side of the figure is comparison of branch fusion accuracy between 8 times and 32 times of resolution reduction, and on the right hand side of the figure is Operation speed}
\label{fig8}
\end{figure}

\begin{figure}[!t]
\centering
\includegraphics[width=\columnwidth]{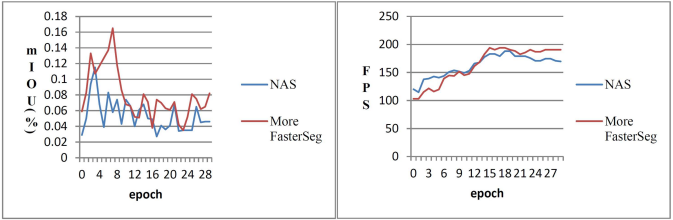}
\caption{Use the Cityscapes data set to use neural search technology to generate the student network in the knowledge distillation network.
On the left hand side of the figure is 8-fold decrease in resolution and accuracy of 32-fold fusion branched comparison chart, and on the right hand side of the figure is Operation speed}
\label{fig9}
\end{figure}
 
 From the analysis of Figure 8, it is found that the accuracy of the teacher network model generated by adding an adaptive attention mechanism to the output branch of the teacher network model in the neural network architecture search technology proposed in this paper is not as good as that of the NAS technology. The speed of processing images is higher than that of NAS technology. Comparing Figure 9 (a) and Figure 9 (b), it is found that the student network model generated by adding an adaptive attention mechanism to the output branch of the student network model in the neural network architecture search technology is higher than NAS technology in terms of accuracy and processing speed.
 
 The analysis based on the above experimental data is as follows: adding an adaptive attention module to the architecture search technology is not very useful for the teacher network model, but for the student network model, it has a greater improvement in accuracy and processing speed. The key reason is that the teacher network in the search process of the search technology does not limit the search expansion ratio, which makes the teacher network model more complicated than the student network model. After the adaptive attention module is added, the information of the redundant feature channel will be removed, which will lead to the teacher network. The complexity of the model decreases, which reduces the accuracy of the teacher network model. For the student network model, after using the adaptive attention module to remove redundant feature channel information, the processing speed is accelerated. Under the condition of limited complexity, the effective pixel feature information is retained more. Therefore, the adaptive attention module is added to the output branch of the student network model. The attention mechanism module is effective.

\subsubsection{Evaluation of the accuracy and complexity of the teacher network model}
This section takes the test set and verification set of Cityscapes as the experimental objects to further evaluate the accuracy and complexity of the teacher network model. The data comparison shown in Figure 2 shows that in the neural network architecture search technology combined with supernet, the processing speed of the teacher network model branch is the best when the adaptive attention module is added. Since mobile robot navigation has high requirements for the speed of the semantic segmentation process, this paper selects the search technology of adding an adaptive attention module to the teacher network model to generate the M-FasterSeg teacher network model.
Next, an experiment is conducted on whether to add an adaptive attention mechanism module as a post-processing module in the teacher network model. The result data is shown in Table 1. In Table 1, the More FasterSeg* teacher network model uses the results of training without an adaptive attention network module, while the More FasterSeg teacher network model uses the results of training with an adaptive attention mechanism module.

\begin{table}[]
\caption{Teacher network accuracy}
\centering
\setlength{\tabcolsep}{0.6mm}{
\begin{tabular}{cccc}
\hline
Settings                       & mIoU(\%)             & Flops                & parameters           \\ \hline
More FastSeg(T)(Non-Attention) & 74.4                 & 26.38G               & 3.09M                \\ \hline
More FasterSeg*(T)(Attention)  & 75.7                 & 26.15G               & 2.90M                \\ \hline
\multicolumn{1}{l}{}           & \multicolumn{1}{l}{} & \multicolumn{1}{l}{} & \multicolumn{1}{l}{}
\end{tabular}
}
\end{table}

It can be seen from Table 1 that under the image input of the same data set, the teacher network model with the adaptive attention mechanism has higher accuracy, lower floating point calculations and lower parameters. This shows that the adaptive attention module designed in this paper can still achieve better accuracy while reducing the computational complexity of the model. Adding the adaptive attention module as a post-processing module can effectively reduce the redundant information and make the network model more efficient. The semantic segmentation diagrams obtained from the two teacher network models are shown in Figure 10.
\begin{figure}[!t]
\centering
\includegraphics[width=\columnwidth]{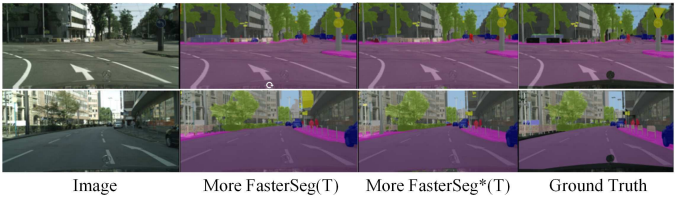}
\caption{Comparison of Semantic Segmentation Results of Teacher Network.
}
\label{fig10}
\end{figure}
It can be seen from Figure 4 that the semantic segmentation output of the teacher network model with the adaptive attention module is higher, especially the target edge is better. In summary, adding an adaptive attention network module to the teacher network model as a post-processing module helps improve the accuracy of the teacher network model, while reducing the amount of invalid calculations of the teacher network model and reducing the consumption of training resources. The research results have played a good role in enhancing the scene understanding ability of mobile robot navigation.

\subsubsection{Accuracy and efficiency evaluation of student network model}
This section will compare the performance of the semantic segmentation method designed in this article with the current representative methods. In addition, since the goal of the method in this paper is to apply to mobile robots, image preprocessing methods such as multi-scale, cleaning data, and flipping are not used in the experiment.

In the process of testing and verifying the student network model, the test set and verification set of Cityscapes are used as experimental objects. From the comparison of the experimental data in Figure 3, it can be seen that in the neural network architecture search technology combined with supernet, the student network model output branch adds an adaptive attention module, and the accuracy and processing speed of the generated network model are optimal, so the student network model is selected As the algorithm model of M- FasterSeg.

This paper focuses on whether to add an adaptive attention mechanism module as a post-processing module in the student network model. The resulting semantic segmentation result map, as shown in Figure 11, has a good segmentation effect on the edge of the object.

\begin{figure}[!t]
\centering
\includegraphics[width=\columnwidth]{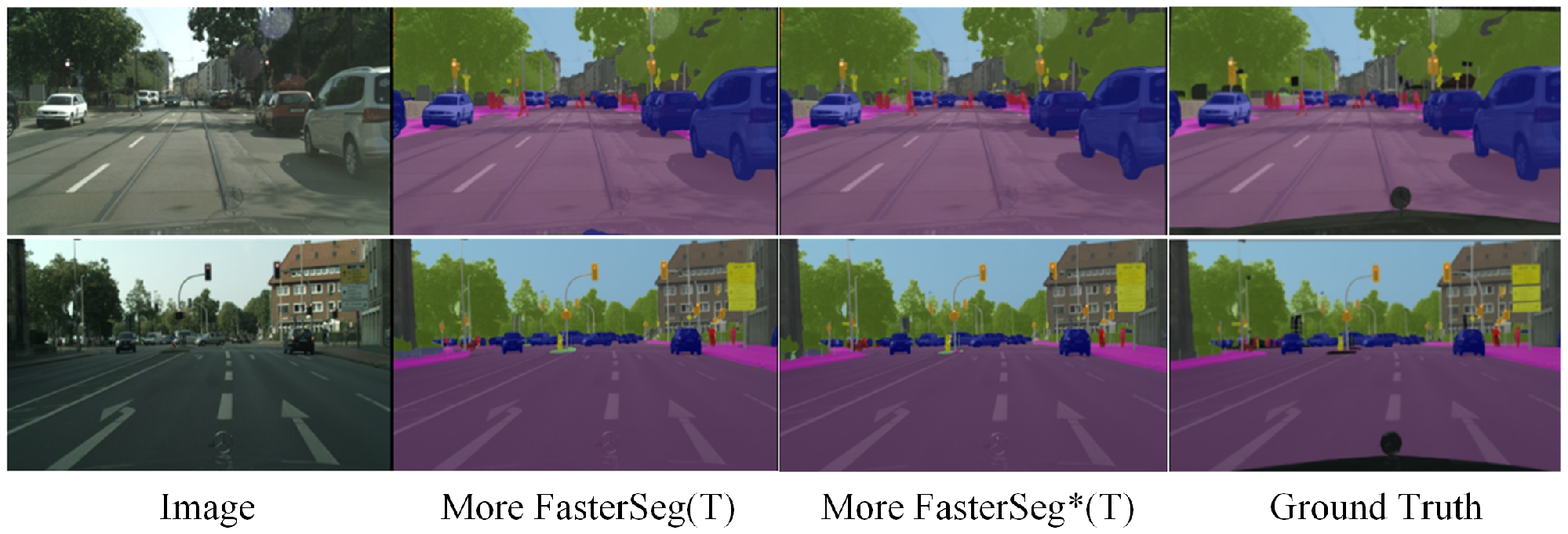}
\caption{Comparison of semantic segmentation results.
}
\label{fig11}
\end{figure}

The experiment uses the test set and validation set of Cityscapes to evaluate the effectiveness of the semantic segmentation network model, and compares the semantic segmentation accuracy of the student network model with the current excellent methods. As shown in Table 2, the experimental results show that the semantic segmentation method proposed in this paper has better accuracy on the verification set and test set, and the calculation efficiency is higher.
\begin{table*}[]
\caption{Real-time semantic segmentation performance of adaptive attention network in Cityscapes dataset}
\centering
\setlength{\tabcolsep}{3.2mm}{
\begin{tabular}{ccccc}
\hline
Method          &Val MIOU(\%) & Test MIOU(\%) & FPS   & Resolution \\ \hline
FCN-8S          & -            & 65.3          & 4.4   & 512x1024   \\
DeepLabv3       & -            & 81.3          & 1.3   & 769x769    \\
ENet            & -            & 58.3          & 76.9  & 512x1024   \\
ICNet           & 67.7         & 69.1          & 37.7  & 1024x2048  \\
BiseNet         & 69.0         & 68.4          & 105.8 & 768x1536   \\
DFANet          & -            & 67.1          & 120   & 1024x2048  \\
Fast-SCNN       & 68.6         & 68.0          & 123.5 & 1024x2048  \\
ESPNetV2        & 66.4         & 66.3          & -     & 1024x2048  \\
DABNet          & -            & 70.1          & 70.1  & 1024x2048  \\
More FasterSeg  & 71.5         & 69.2          & 164.3 & 1024x2048  \\ \hline
More FasterSeg* & 72.1         & 69.8          & 166.4 & 1024x2048  \\ \hline
\end{tabular}}
\end{table*}

In Table 2, the More FasterSeg* student network model uses the results of training without an adaptive attention network module, and the More FasterSeg student network structure model uses the results of training with an adaptive attention mechanism module. The fast semantic segmentation model proposed in this paper is the More FasterSeg model, which compares the semantic segmentation results of the existing excellent semantic segmentation network models on the Cityscapes dataset, as shown in Figure 11.

\begin{figure}[!t]
\centering
\includegraphics[width=\columnwidth]{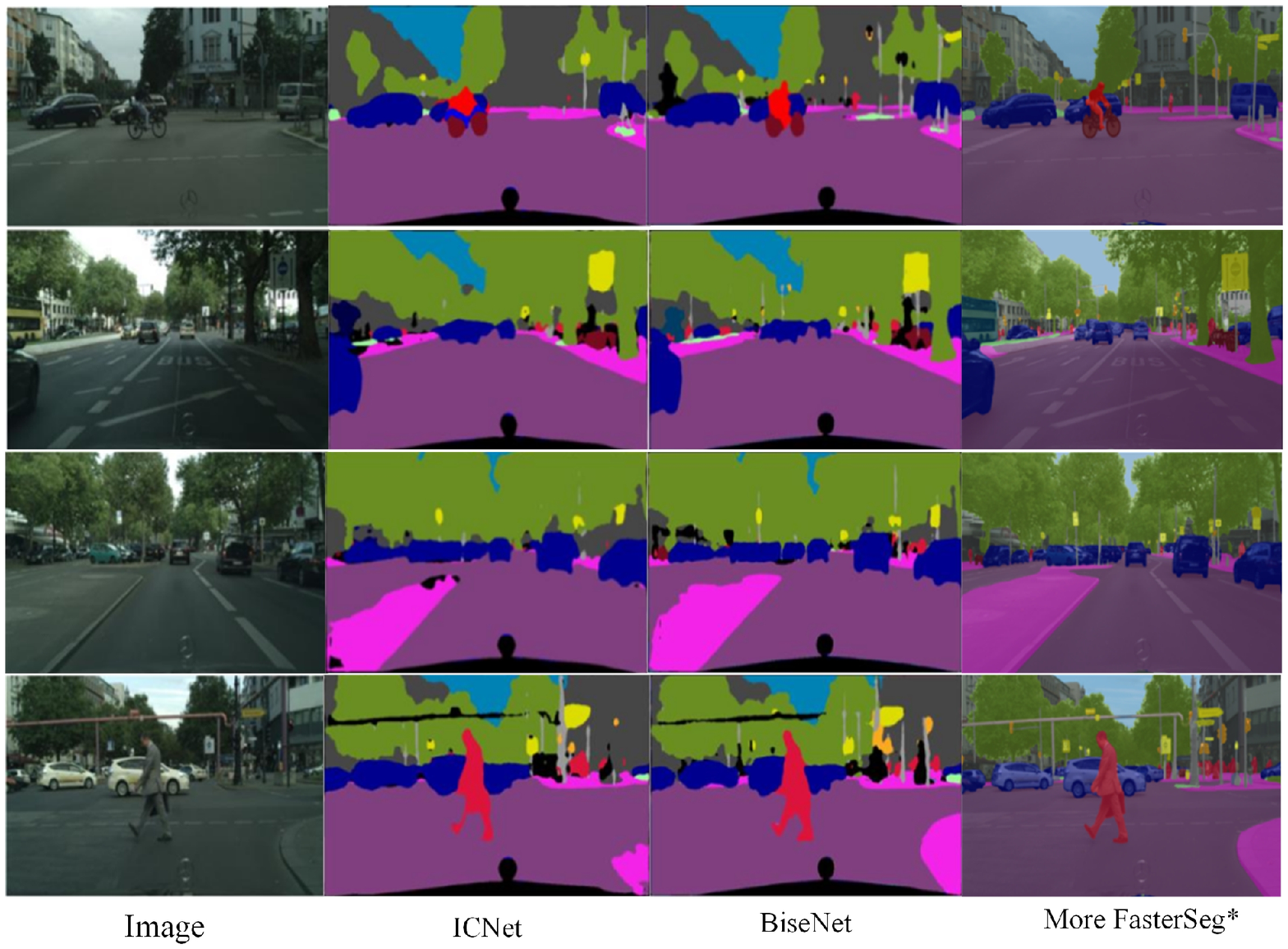}
\caption{Comparison of semantic segmentation results.
}
\label{fig12}
\end{figure}
It can be seen from Figure 11: It is intuitively shown that on the road scene data set, the method in this paper is more effective than the current representative semantic segmentation network. Combined with Table 2, it can be seen that the method in this paper has better real-time performance while achieving good accuracy, which has practical application value for the development of mobile robots’ understanding of complex road scenes and map construction.

\section{Conclusion}
This paper proposes a real-time image semantic segmentation network based on neural network architecture search technology combined with adaptive attention module. The network first optimizes neural network architecture search technology and integrates adaptive attention module and neural network architecture search technology. The output neural network architecture model under the multi-resolution branch can be better optimized. The use of Supernet technology enables the neural architecture search technology to search for student and teacher networks under the same network framework, and uses knowledge distillation technology to combine the teacher network and the student network Combined, the teacher network will not limit the expansion ratio during the search process, while the student network will control its expansion ratio to avoid affecting the real-time performance, which will make the teacher network learn more data during the training process. After the teacher network training is completed, the weight parameters of the teacher network training are used to guide the training of the student network using a method similar to transfer learning. After the student network is affected by the teacher network, the result of semantic segmentation can be obtained in a shorter time. Finally, a series of comparative experiments were carried out on the Cityscapes data set. The experimental results show that the algorithm in this paper can improve the problem of object category edge loss, and can more accurately deal with road scene images in the case of multi-object aliasing. Image semantic segmentation problem. At the same time, it is proved that the algorithm in this paper is a real-time image segmentation algorithm that can effectively take into account real-time and accuracy.



 \bibliographystyle{elsarticle-harv} 
 \bibliography{mybibliography}





\end{document}